\documentclass{article}

    \PassOptionsToPackage{numbers, compress}{natbib}

    \usepackage[final]{neurips_2020}

\usepackage[utf8]{inputenc} 
\usepackage[T1]{fontenc}    
\usepackage{hyperref}       
\usepackage{url}            
\usepackage{booktabs}       
\usepackage{amsfonts}       
\usepackage{nicefrac}       
\usepackage{microtype}      

\usepackage{subcaption}
\usepackage{graphicx}
\usepackage{algorithmic}
\usepackage{float}
\usepackage{amsmath}
\title{Front Contribution instead of Back Propagation}

\author{%
  Swaroop Mishra \\
    Arizona State University\\
    \texttt{srmishr1@asu.edu}
   \And
   Anjana Arunkumar \\
   Arizona State University \\
    \texttt{aarunku5@asu.edu}
}

\begin{document}

\maketitle

\begin{abstract}
Deep Learning's outstanding track record across several domains has stemmed from the use of error backpropagation (BP). Several studies, however, have shown that it is impossible to execute BP in a real brain. Also, BP still serves as an important and unsolved bottleneck for memory usage and speed. We propose a simple, novel algorithm, the Front-Contribution algorithm, as a compact alternative to BP. The contributions of all weights with respect to the final layer weights are calculated before training commences and all the contributions are appended to weights of the final layer, i.e., the effective final layer weights are a non-linear function of themselves. Our algorithm then essentially collapses the network, precluding the necessity for weight updation of all weights not in the final layer. 
This reduction in parameters results in lower memory usage and higher training speed. We show that our algorithm produces the exact same output as BP, in contrast to several recently proposed algorithms approximating BP. Our preliminary experiments demonstrate the efficacy of the proposed algorithm. Our work provides a foundation to effectively utilize these presently under-explored "front contributions", and serves to inspire the next generation of training algorithms.
\end{abstract}
\section{Introduction and Related Work}
Backpropagation of error (BP) \cite{rumelhart1985learning} has been the best algorithm to train neural networks, and has driven deep learning to perform outstandingly across several domains \cite{lecun2015deep}. However, it is not consistent with our findings about the brain \cite{crick1989recent,izhikevich2008large}. In fact, it is not possible to execute BP in a real brain \cite{bartunov2018assessing}. BP also suffers from several other problems, such as a vanishing/exploding gradient. Inspite of using careful initialization and architecture modifications \cite{ororbia2018conducting, glorot2010understanding}-- for example, using RELU instead of sigmoid activations-- quite effectively as workarounds, BP still is a key bottleneck for memory usage and speed. This may indicate that \textit{BP is a suboptimal algorithm and will be replaced}.


Several different algorithms have been proposed to improve BP. However,
those algorithms have tried to approximate BP and have not been able to work beyond toy datasets \cite{bartunov2018assessing}; thus they cannot be applied in a real world setting. We survey three categories of BP literature-- (i) better hardware implementation of BP \cite{kumar2019efficient, kusumoto2019graph,wangbackpropagation, gruslys2016memory, xu2018backprop, ororbia2020reducing}, (ii) workarounds to approximate BP \cite{zhang2019spike, ernoult2019updates, gomez2017reversible}, and (iii) biologically inspired algorithms. Biologically inspired algorithms can further be segregated into four types: (i) Inspired from biological observations \cite{sacramento2018dendritic, ernoult2019updates, ororbia2019biologically, lansdell2019learning},  these works try to approximate BP with the intention resolve its biological implausibility, (ii) Propagation of an alternative to error \cite{lee2015difference, manchev2020target}, (iii) Leveraging local errors, the power of single layer networks, and layer wise pre-training to approximate BP
\cite{nokland2019training, mostafa2018deep, belilovsky2018greedy}, (iv) Resolving the locking problem using decoupling \cite{jaderberg2017decoupled, czarnecki2017understanding, huo2018decoupled, baldi2016learning, ma2019hsic} and its variants \cite{ororbia2018conducting, flennerhag2018breaking, miyato2017synthetic, choromanska2018beyond}. We were deeply motivated by (ii), (iii), and (iv) while coming up with the idea of `front contributions'-- specifically, propagating something other than error, the idea of a single layer network, and decoupling, collectively inspire `front contributions'. The key distinction of our front contribution is that, it produces exact same output as BP unlike other approaches that have tried to approximate BP.

\section{Method}


We know that a set of linear layers can be collapsed to a single layer network; however, non-linear activation functions have hitherto restricted the collapse of networks that they are applied in. Here, we justify that every fully connected multi-layer network can be collapsed to a single layer network, using the \textit{Front Contribution Algorithm}, eliminating the requirement of backpropagation \footnote{See Supplementary Material: Analogy for illustrative explanation of backpropagation and front contribution}. 

\textbf{Formalization:}

Let a neural network (NN) have $n$ layers,  such that each layer of its weights are represented as $W_i : i\epsilon[1,n]$, where $W_1$ is applied to the input and $W_n$ is on the branch connected to the output node. Let the input of NN be $X$, intermediate layer outputs be $v_j : j\epsilon[1,n-1]$, and final output be $Y$. We can represent $Y$ as:
\begin{equation}
Y=f(X, W_1,W_2.....W_n)
\label{deq1}
\end{equation}

Here, each of the weights are randomly initialized, and then updated using conventional back propagation. Using the chain rule, we have:

\begin{align}
    \Delta W_1=f(W_2,W_3...W_n)\label{deq2}\\
    \Delta W_2=f(W_3,W_4...W_n)\label{deq3}
\end{align}

From equation \ref{deq1}, we see that the input $X$ is multiplied with the weights, and then transformed by activations at various layers to produce the output $Y$. In other words, we can say that weight layers $W_1...W_n$ indirectly contribute towards deciding what $Y$ will be for a given $X$, in a hierarchical sequence starting from $W_1$ till $W_n$. The definition of `contribution' varies depending on the application-- for example, in the case of language models like BERT, contribution refers to attention\cite{Vaswani2017AttentionIA}.

Now, from equations \ref{deq2},\ref{deq3} we see that in the training process, for the $i$\textsuperscript{th} layer, $\Delta$ W is a function of all the $W_i$ layers, from the $(i+1)$\textsuperscript{th} layer up till the $n$\textsuperscript{th} layer. From equation \ref{deq2}, we see that $\Delta$ $W_1$ is a function of $W_2...W_n$, but does not further depend on $W_1$, i.e., the value of $W_1$ at any iteration of training depends on the initial value of $W_1$-- a random static value-- and weights $W_2...W_n$. So, we can say that $W_1$ is not part of the system basis-- the set of vectors that can be used to represent any system state, such as the use of $x$, $y$, $z$ bases to represent any state of a 3D system-- as $W_2...W_n$ can represent the output value at any iteration. So, vector $W_1$ is not actually a necessary variable.

For example, by defining a 3D system in terms of $x$, $y$, $z$, $x+y+z$, we utilize an unnecessary variable, $x+y+z$. As $W_1$ is not an independent weight layer, we are therefore wasting GPU space by using it. However, if the weight layer $W_1$ is ignored, will the system still function as desired? The value of $v_1$ depends on weights in $W_1$, so if $W_1$ weights are not updated, $v_2$ will retain its old, incorrect value, consequently affecting $Y$ even if all other weights are updated correctly, as the network is connected as a hierarchy.

How do we compensate the non-updation of weights in $W_1$? Let a compensation weight $p$ be added to weights in $W_2$ such that $Y$ remains the same, i.e., $p$ compensates for the change in $v_2$ that normally happens with the updation of $W_1$. Here, $p$ must be a non-linear function of weights in $W_2$, as it compensates for an update that depends on  $W_2...W_n$. $p$ can be derived by equating the $v_2$ value found post updation of $W_1$ in conventional backpropgation, and the $v_2$ value calculated without updating $W_1$ and replacing weights in $W_2$ as $W_2+p$.

\section{Finding an Expression for $p$:}

\begin{figure}
    \centering
    \includegraphics[width=\textwidth]{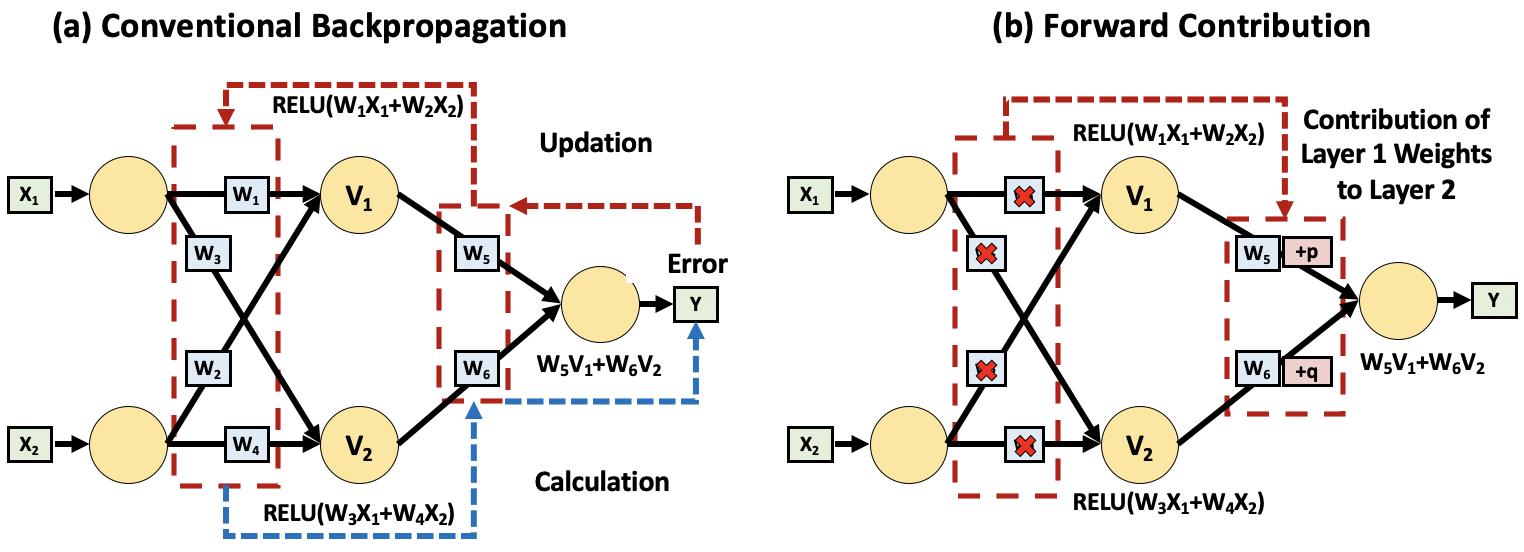}
    \caption{Neural network structure for (a) Backpropagation, (b) Forward Contribution. Here, layer $W_1$ contains weights $w_1$, $w_2$, $w_3$, $w_4$ and $W_2$ contains weights $w_5$, $w_6$.}
    \label{fig:nn}
    \vspace{-5mm}
\end{figure}

 Given a 2 layer neural network as shown in Figure \ref{fig:nn}, where inputs are $x_1$, $x_2$, and a single output $Y$ is produced, if $w_1,w_2,w_3,w_4 \epsilon W_1$ are not updated, then non-linear `compensation weights' of $p$, $q$ must be added to $w_5,w_6 \epsilon W_2$ respectively in order to preserve $Y$ output updation. Here, we consider outputs at intermediate layer nodes after RELU activation to be $v_1, v_2$ (for example, $s_1=w_1x_1+w_2x_2$; $v_1=RELU(s_1)$).
 
To derive values\footnote{Refer to Supplementary Material for the full derivation} of $p,q$, consider that for the initial updation, we can say that:
\begin{align}
Y_{backpropgation}&=Y_{contributionfactor}
\implies v_1.w_5+v_2.w_6&=v_1^c.(w_5+p)+v_2^c.(w_6+q)
\label{eq:13}
\end{align}
where $v_1^c$ and $v_2^c$ are the constants-- their values don't change, as $w_1$,$w_2$,$w_3$,$w_4$ don't get updated.

During updation between the 0\textsuperscript{th} and 1\textsuperscript{st} iterations, $dv_1$' can have 3 possible values based on active and dead RELU before and after updation.


 Consider a single input scheme, i.e., [$x_1$,$x_2$], such that additional compensation to $w_5$ at each iteration is:\\
\begin{align*}
\hspace{0.3cm} {\frac{dE}{dw_5'}}=p
\vspace{-4mm}
\end{align*}

Using the chain rule, we can generalize this to calculate the total compensation weight $P$ that must be added at the $n_{th}$ iteration. Here, $p$ and $P$ compensate for weights $w_1,w_2$. Similarly, $q$ and $Q$ can be defined to compensate for weights $w_3,w_4$.  Let $w_5+p$ be $r_5$ and $w_6+q$ be $r_6$. $r$ may be interpreted as a transformation on $W_2$ that takes care of the contribution factor from $W_1$. Hence, the general equation for updation after `n' iterations is:

\begin{align*}
 r_{5n}&=w_{5n}+\frac{w_{5n}^3(x_1^2+x_2^2)}{3(v_{1c}+\sum_1^n \frac{dE}{dY}_{n} w_{5n}(x_1^2+x_2^2)^2)}\hspace{0.2cm} (A_{n-1}>0\hspace{0.2cm}\&\& \hspace{0.2cm}A_{n}>0)\\
   &=w_{5n}+\frac{w_{5n}^2}{2\eta\frac{dE}{dY}_{n}(v_{1c}+\sum_1^n \frac{dE}{dY}_{n} w_{5n}(x_1^2+x_2^2))}\hspace{0.2cm}(A_{n-1}>0\hspace{0.2cm}\&\&\hspace{0.2cm}A_{n}<0)\\
\end{align*}

where $A_n=(w_{1c}+\sum_1^n \frac{dE}{dY}_{n} w_{5n}x_1)x_1+(w_{2c}+\sum_1^n \frac{dE}{dY}_{n} w_{5n}x_2)x_2$\\

\begin{align*}
 r_{6n} &=w_{6n}+\frac{w_{6n}^3(x_1^2+x_2^2)}{3(v_{2c}+\sum_1^n \frac{dE}{dY}_{n} w_{6n}(x_1^2+x_2^2)^2)}\hspace{0.2cm} (B_{n-1}>0\hspace{0.2cm}\&\&\hspace{0.2cm}B_{n}>0)\\
   &=w_{6n}+\frac{w_{6n}^2}{2\eta\frac{dE}{dY}_{n}(v_{2c}+\sum_1^n \frac{dE}{dY}_{n} w_{6n}(x_1^2+x_2^2))}\hspace{0.2cm}(B_{n-1}>0\hspace{0.2cm}\&\&\hspace{0.2cm}B_{n}<0)\\
\end{align*}

where $B_n=(w_{3c}+\sum_1^n  \frac{dE}{dY}_n w_{6n}x_1)x_1+(w_{4c}+\sum_1^n  \frac{dE}{dY}_n w_{6n}x_2)x_2$\\

Here subscript $n$ represents value at the $n^{th}$ iteration, and subscript $c$ represents the constant value, i.e., the initial value which does not change across iterations.\\

We find that for multiple inputs, the entire input sequence must be stored to calculate the $P$ and $Q$ values. Therefore, in the cases of updating either (i) $w_5,w_6$ by adding compensatory weights $p,q$, or (ii) updating the compensatory weights $p,q$ for a fixed $w_5, w_6$, we use a single input to run a basic exploratory analysis experiment\footnote{Refer to Supplementary Materials: Experiments}.

To handle multiple inputs, we update $s_1$, $s_2$ as follows:

\begin{align}
\text{Initially: }
    s_1=w_1x_1+w_2x_2,\ s_2=w_3x_1+w_4x_2,
    S_1=w_1x_1-w_2x_2,\ S_2=w_3x_1-w_4x_2\nonumber\\
\text{Updated as: }
    s_1\textsuperscript{@}'=\frac{s_1'+S_1'}{2}x_1\textsuperscript{@}+\frac{s_1'-S_1'}{2}x_2\textsuperscript{@},\
    s_2\textsuperscript{@}'=\frac{s_2'+S_2'}{2}x_1\textsuperscript{@}+\frac{s_2'-S_2'}{2}x_2\textsuperscript{@}\nonumber\\
    S_1\textsuperscript{@}'=\frac{s_1'+S_1'}{2}x_1\textsuperscript{@}-\frac{s_1'-S_1'}{2}x_2\textsuperscript{@},\
    S_2\textsuperscript{@}'=\frac{s_2'+S_2'}{2}x_1\textsuperscript{@}-\frac{s_2'-S_2'}{2}x_2\textsuperscript{@}
\end{align}

Above, we have shown that in the context of \ref{deq1}, for weight layer $W_2$, ($W_2+p$) can contribute (attend for text) equivalent to updated $W_1$, and $W_2$. $p$ is the contribution/weight attention factor of $W_1$ to $W_2$, i.e., the amount by which $W_1$ contributes/attends to $W_2$. In this case, $W_2$ weights are now non-linear, and we find that $W_2$ weights can take the form $W_2+W_2^3$. By introducing non-linear weights, we can therefore reduce the total number of weights by 1 set (i.e., the number of weights corresponding to a particular layer). By showing that NN of n layers can be collapsed to n-1 layers, we can further use the method of induction to theoretically prove that NN can be collapsed from $(n-1)->(n-2)->...->1$ layer, such that the final network has just non-linear weights for layer $W_n$.

This method of $W_n$ calculation is the \textit{Front Contribution Algorithm}, as instead of propagating error backwards, the network propagates contribution forward to collapse the network. 
\section{Conclusion:}
We proposed a simple, novel algorithm, the Front-Contribution algorithm, as a compact alternative to BP. Our algorithm has several advantages:

(i) Front Contribution is a one-time calculation, i.e., given a NN structure, with number of layers, activations, etc., the algorithm will output the non-linear weights of a corresponding collapsed network. While the expression for $W_n$ weights might be complex, the algorithm's time complexity remains $O(1)$.\\
(ii) As the number of weights in a collapsed network is much lesser than in the original network, GPU space usage will be drastically improved. Deep Learning has significantly benefited from parallel computing, so by substantially improving parallel computing performance, we can potentially create deeper networks.\\
(iii) Due to the absence of back-propagation, we expect network training time to also be drastically reduced.\\
(iv) The post-training non-linear weights (represented in terms of their layers, $W_i$)-- consider $W_n=W_n+k_1W_n^3+k_2W_n^5+...$-- can potentially represent the features that the network has learned. Later terms in the preceding expression should represent simpler features (like edges), while initial terms should represent more complex features. However, $W_n$ will give us an idea of the aggregate features that are important for NN. We define aggregate features as the final set of features that NN cares about, after attending to various features in a hierarchical manner.\\
(v) The modified network can always be expanded to a conventional network depending on our requirements-- for example, we might want to see activations of attention neurons, and can therefore re-expand the network up till those neurons, without having to expand it fully.\\
(vi) This derivation can be modified and extended to other types of networks, such as CNNs, RNNs and transformers.

\bibliography{neurips_2020.bib}
\bibliographystyle{abbrvnat}

\section{Supplemental Material}
\label{sec:supplemental}
We know that a set of linear layers can be collapsed to a single layer network; however, non-linear activation functions have hitherto restricted the collapse of networks that they are applied in. Here, we justify that every fully connected multi-layer network can be collapsed to a single layer network, using the \textit{Front Contribution Algorithm}, eliminating the requirement of backpropagation. 

\textbf{Analogy:}

Consider an assembly line, where 10 people ($P_1-P_{10}$) work sequentially to produce a toy car from  plastic. $P_1$ cuts raw plastic (input) into different sizes, and gives it to $P_2$, $P_2$ smooths the plastic and gives it to $P_3$ (intermediate output), and so on until $P_{10}$ produces the toy car (final output). Initially, all the people do the task with no prior knowledge, so they perform random actions, aimed at producing the final output (car). $P_{10}$ sees the output produced, compares it to the desired toy car, and rectifies his actions based on the input he receives from $P_9$, to make the output closer and closer to the toy car. $P_9$, $P_8$, etc. do the same-- they look at the original error that they receive, find out how much they contribute to the error, and then try to rectify their actions by small amounts (not knowing exactly how much to rectify) to produce a perfect toy car. Therefore, every individual finds their respective individual actions to be performed and collectively produce a toy car. They can use this approach to build many other things, such as houses, fountains, etc., until they succeed as often as possible. Is this however an efficient method of learning to build anything?

Consider a person, $P_{11}$, who observes that $P_2$ is merely doing a fine-tuned version of $P_1$'s work by getting plastic to a desired physical state, and can therefore reliably do both $P_1$'s and their own work as they both are new to building. If $P_1$ is eliminated, and $P_2$ is asked to both cut and smooth plastic, by hiring one less person and reducing cost of transfer between $P_1$ and $P_2$, the budget allows for extra training time for $P_2$. $P_2$ also doesn't need to necessarily take more time to complete their work, if given a tool that both cuts and smooths the cut edges of plastic simultaneously-- i.e., the output from $P_2$ becomes the sum of rectified effort from $P_1$ and $P_2$. This is a one time cost, and $P_{11}$ can develop the required tool by finding the `contribution' of $P_1$'s actions to $P_2$, i.e., how much $P_1$'s work attends to $P_2$. Similarly, by observing the contributions of workers down the assembly line, $P_{11}$ is able to eliminate all but the last worker, and provide him with a single tool that can be used to perform all the work. This results in a drastic decrease in budget and increase in building speed. It also allows more time for process and output analysis, to understand how different steps and features contribute towards a specific task and overall output-- the process becomes more transparent, akin to process explainability.

This can be extended towards other concepts such as center of mass in physics as well. We aim to calculate a single-layer network with non-linear weights (calculated with Forward Contribution) whose performance is tantamount to a conventional multi-layer network that uses backpropagation. This is not an approximation (mimic) network, but an equivalent network to BP.

\textbf{Formalization:}

Let a neural network (NN) have $n$ layers,  such that each layer of its weights are represented as $W_i : i\epsilon[1,n]$, where $W_1$ is applied to the input and $W_n$ is on the branch connected to the output node. Let the input of NN be $X$, intermediate layer outputs be $v_j : j\epsilon[1,n-1]$, and final output be $Y$. We can represent $Y$ as:
\begin{equation}
Y=f(X, W_1,W_2.....W_n)
\label{deq1}
\end{equation}

Here, each of the weights are randomly initialized, and then updated using conventional back propagation. Using the chain rule, we have:

\begin{align}
    \Delta W_1=f(W_2,W_3...W_n)\label{deq2}\\
    \Delta W_2=f(W_3,W_4...W_n)\label{deq3}
\end{align}

From equation \ref{deq1}, we see that the input $X$ is multiplied with the weights, and then transformed by activations at various layers to produce the output $Y$. In other words, we can say that weight layers $W_1...W_n$ indirectly contribute towards deciding what $Y$ will be for a given $X$, in a hierarchical sequence starting from $W_1$ till $W_n$. The definition of `contribution' varies depending on the application-- for example, in the case of language models like BERT, contribution refers to attention.

Now, from equations \ref{deq2},\ref{deq3} we see that in the training process, for the $i$\textsuperscript{th} layer, $\Delta$ W is a function of all the $W_i$ layers, from the $(i+1)$\textsuperscript{th} layer up till the $n$\textsuperscript{th} layer. From equation \ref{deq2}, we see that $\Delta$ $W_1$ is a function of $W_2...W_n$, but does not further depend on $W_1$, i.e., the value of $W_1$ at any iteration of training depends on the initial value of $W_1$-- a random static value-- and weights $W_2...W_n$. So, we can say that $W_1$ is not part of the system basis-- the set of vectors that can be used to represent any system state, such as the use of $x$, $y$, $z$ bases to represent any state of a 3D system-- as $W_2...W_n$ can represent the output value at any iteration. So, vector $W_1$ is not actually a necessary variable.

Since $W_1$ is not part of the basis, it is actually not a necessary variable; for example, by defining a 3D system in terms of $x$, $y$, $z$, $x+y+z$, we utilize an unnecessary variable, $x+y+z$. As $W_1$ is not an independent weight layer, we are therefore wasting GPU space by using it. However, if the weight layer $W_1$ is ignored, will the system still function as desired? The value of $v_1$ depends on weights in $W_1$, so if $W_1$ weights are not updated, $v_2$ will retain its old, incorrect value, consequently affecting $Y$ even if all other weights are updated correctly, as the network is connected as a hierarchy.

How do we compensate the non-updation of weights in $W_1$? Let a compensation weight $p$ be added to weights in $W_2$ such that $Y$ remains the same, i.e., $p$ compensates for the change in $v_2$ that normally happens with the updation of $W_1$. Here, $p$ must be a non-linear function of weights in $W_2$, as it compensates for an update that depends on  $W_2...W_n$. $p$ can be derived by equating the $v_2$ value found post updation of $W_1$ in conventional backpropgation, and the $v_2$ value calculated without updating $W_1$ and replacing weights in $W_2$ as $W_2+p$.

\section{Finding an Expression for \textit{p}}

\begin{figure}
    \centering
    \includegraphics[width=\textwidth]{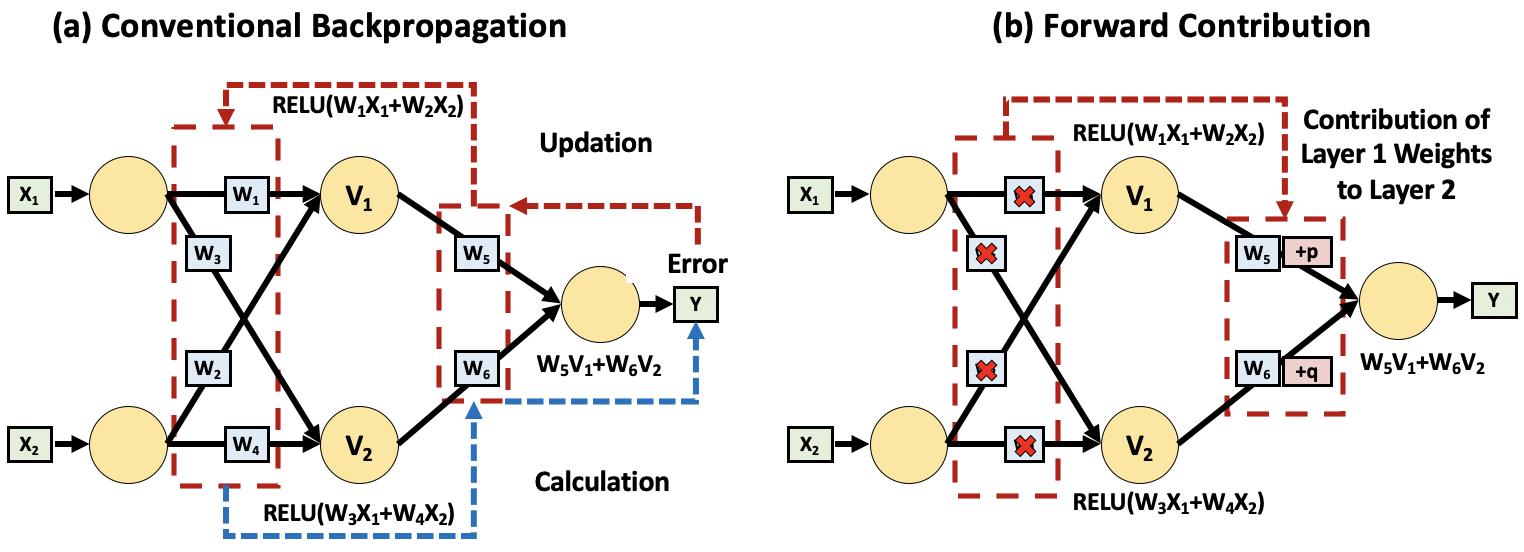}
    \caption{Neural network structure for (a) Backpropagation, (b) Forward Contribution. Here, layer $W_1$ contains weights $w_1$, $w_2$, $w_3$, $w_4$ and $W_2$ contains weights $w_5$, $w_6$.}
    \label{fig:nn}
\end{figure}
Let $X$ be the input vector: [[$x_{1}$,$x_{2}$], [$x_{1}$\textsuperscript{@},$x_{2}$\textsuperscript{@}], [$x_{1}$\textsuperscript{*},$x_{2}$\textsuperscript{*}], [$x_{1}$\textsuperscript{\#}, $x_{2}$\textsuperscript{\#}]]. Let $Y_{g}$ be the vector of gold labels: [$y_{g}$, $y_{g}$\textsuperscript{@}, $y_{g}$\textsuperscript{*}, $y_{g}$\textsuperscript{\#}]. Let $w_{1}$, $w_{2}$, $w_{3}$, $w_{4}$ be the weights of the first layer ($W_1$). Let $w_{5}$, $w_{6}$ be the weights of the last layer ($W_2$). We define the state of the intermediate nodes $v_{1}$ and $v_{2}$ for a given input as:
\begin{equation}
    s_{1}=w_{1} \cdot x_{1} + w_{2} \cdot x_{2}, 
    s_{2}=w_{3} \cdot x_{1} + w_{4} \cdot x_{2}\\
    \label{eq:1}
\end{equation}
The activation function for intermediate nodes is RELU, i.e.:
\begin{equation}
    v_{1}=RELU(s_{1}), 
    v_{2}=RELU(s_{2})\\\label{eq:2}
\end{equation}
where
    \begin{equation*}
    RELU(s)=
    \begin{cases}
      s &\text{if}\ s>0 \\
      0 & \text{if}\ s<0
    \end{cases}
\end{equation*}
Let $\eta$ be the learning rate and $Y$ be the output vector: [$y$, $y$\textsuperscript{@}, $y$\textsuperscript{*}, $y$\textsuperscript{\#}], where
\begin{equation}
    y=w_{5} \cdot v_{1} + w_{6} \cdot v_{2}
    \label{eq:3}
\end{equation}
Let the error $E$ be defined as:
\begin{equation}
    E=\frac{1}{2}(y-y_{g})\textsuperscript{2} \text{, i.e., } \frac{dE}{dY}=y-y_{g}
    \label{eq:4}
\end{equation}

The following repeats successively for each epoch, for a single given input:
\begin{equation}
w_{5}' = w_{5}-\eta \frac{dE}{dY}v_{1}, 
w_{6}' = w_{6}-\eta \frac{dE}{dY}v_{2}
\label{eq:5}
\end{equation}

$\frac{dE}{dw_1}$ can have 2 possible values based on whether the RELU is dead or not, as can $\frac{dE}{dw_2}$,$\frac{dE}{dw_3}$ and $\frac{dE}{dw_4}$:
    \begin{align*}
        \frac{dE}{dw_1}=0\ \text{(or)}\
        \frac{dE}{dw_1}=\frac{dE}{dY}.w_5.x_1(s_1>0)\\
        \frac{dE}{dw_2}=0\ \text{(or)}\
        \frac{dE}{dw_2}=\frac{dE}{dY}.w_5.x_2(s_1>0)\\
        \frac{dE}{dw_3}=0\ \text{(or)}\
        \frac{dE}{dw_3}=\frac{dE}{dY}.w_6.x_1(s_2>0)\\
        \frac{dE}{dw_4}=0\ \text{(or)}\
        \frac{dE}{dw_4}=\frac{dE}{dY}.w_6.x_2(s_2>0)\\
    \end{align*}

Based on this, we update the weights of the first layer as:
\begin{align}
w_{1}' = w_{1}-\eta \frac{dE}{dY} w_{5}'x_{1} (s_1>0) \text{(or)} w_{1}' = w_{1} (s_1<0)
\label{eq:6}
\end{align}
\begin{align*}
w_{2}' = w_{2}-\eta \frac{dE}{dY} w_{5}'x_{2} (s_1>0) \text{(or)} w_{2}' = w_{2} (s_1<0)
\end{align*}
\begin{align*}
w_{3}' = w_{3}-\eta \frac{dE}{dY} w_{6}'x_{1} (s_2>0) \text{(or)} w_{3}' = w_{3} (s_2<0)
\end{align*}
\begin{align*}
w_{4}' = w_{4}-\eta \frac{dE}{dY} w_{6}'x_{2} (s_2>0) \text{(or)} w_{4}' = w_{4} (s_2<0)
\end{align*}

\begin{align}
    s_{1}'=w_{1}' \cdot x_{1} + w_{2}' \cdot x_{2}, 
    s_{2}'=w_{3}' \cdot x_{1} + w_{4}' \cdot x_{2}, 
\label{eq:7}
\end{align}
\begin{align*}
    v_1'=RELU(s_1'), v_2'=RELU(s_2'), y=w_{5}' \cdot v_{1}' + w_{6}' \cdot v_{2}'
\end{align*}

where
\begin{equation}
\frac{dE}{dY}=y-y_{g}=(w_{5} \cdot v_{1} + w_{6} \cdot v_{2})-y_{g} \label{eq:8}
\end{equation}

From equations \ref{eq:6}, \ref{eq:7}:
\begin{align}
s_{1}'=(w_{1}-\eta \frac{dE}{dY} w_{5}'x_{1})x_{1}+(w_{2}-\eta \frac{dE}{dY} w_{5}'x_{2})x_{2}
\label{eq:9}
\end{align}
\begin{align*}
=w_{1} \cdot x_{1} + w_{2} \cdot x_{2}-\eta \frac{dE}{dY} w_{5}'(x_{1}\textsuperscript{2}+x_{2}\textsuperscript{2})
\end{align*}
\begin{align*}
    =s_{1}-\eta \frac{dE}{dY} w_{5}'(x_{1}\textsuperscript{2}+x_{2}\textsuperscript{2})
\end{align*}
\begin{align*}
    \text{Parallely, }s_{2}'=s_{2}-\eta \frac{dE}{dY} w_{6}'(x_{1}\textsuperscript{2}+x_{2}\textsuperscript{2})
\end{align*}

If the next input is [$x_{1}$\textsuperscript{@},$x_{2}$\textsuperscript{@}] then for the next epoch, we have:
\begin{equation}
    s_{1}'\textsuperscript{@}=(w_{1}-\eta \frac{dE}{dY} w_{5}'x_{1})x_{1}\textsuperscript{@} +(w_{2}-\eta \frac{dE}{dY} w_{5}'x_{2})x_{2}\textsuperscript{@}
\label{eq:10}
\end{equation}
\begin{align*}
    =w_{1} \cdot x_{1}\textsuperscript{@} + w_{2} \cdot x_{2}\textsuperscript{@} -\eta \frac{dE}{dY} w_{5}'(x_{1}x_{1}\textsuperscript{@}+x_{2}x_{2}\textsuperscript{@})
\end{align*}
\begin{align*}
    s_{2}'\textsuperscript{@}=(w_{3}-\eta \frac{dE}{dY} w_{6}'x_{1})x_{1}\textsuperscript{@} +(w_{4}-\eta \frac{dE}{dY} w_{6}'x_{2})x_{2}\textsuperscript{@}
\end{align*}
\begin{align*}
    =w_{3} \cdot x_{1}\textsuperscript{@} + w_{4} \cdot x_{2}\textsuperscript{@} -\eta \frac{dE}{dY} w_{6}'(x_{1}x_{1}\textsuperscript{@}+x_{2}x_{2}\textsuperscript{@})
\end{align*}
Using $s_1$\textsuperscript{@}=$w_{1} \cdot x_{1}$\textsuperscript{@} + $w_{2} \cdot x_{2}$\textsuperscript{@}, $s_2$\textsuperscript{@}=$w_{3} \cdot x_{1}$\textsuperscript{@} + $w_{4} \cdot x_{2}$\textsuperscript{@}, and equation \ref{eq:9}, we can simplify equation \ref{eq:10} as:

\begin{align}
s_{1}'\textsuperscript{@}=s_1\textsuperscript{@}+\frac{s_{1}'-s_1}{x_{1}\textsuperscript{2}+x_{2}\textsuperscript{2}}(x_{1}x_{1}\textsuperscript{@}+x_{2}x_{2}\textsuperscript{@})
\label{eq:11}
\end{align}
\begin{align*}
s_{2}'\textsuperscript{@}=s_2\textsuperscript{@}+\frac{s_{2}'-s_2}{x_{1}\textsuperscript{2}+x_{2}\textsuperscript{2}}(x_{1}x_{1}\textsuperscript{@}+x_{2}x_{2}\textsuperscript{@})
\end{align*}

Let:
\begin{align*}
    s_1\textsuperscript{*}=w_{1} \cdot x_{1}\textsuperscript{*} + w_{2} \cdot x_{2}\textsuperscript{*},     s_1\textsuperscript{\#}=w_{1} \cdot x_{1}\textsuperscript{\#} + w_{2} \cdot x_{2}\textsuperscript{\#}
\end{align*}
\begin{align*}
    s_2\textsuperscript{*}=w_{3} \cdot x_{1}\textsuperscript{*} + w_{4} \cdot x_{2}\textsuperscript{*},     s_2\textsuperscript{\#}=w_{3} \cdot x_{1}\textsuperscript{\#} + w_{4} \cdot x_{2}\textsuperscript{\#}
\end{align*}

Using the above with equation \ref{eq:11}, for the next 2 epochs with inputs [$x_{1}$\textsuperscript{*},$x_{2}$\textsuperscript{*}] and [$x_{1}$\textsuperscript{\#},$x_{2}$\textsuperscript{\#}], we get:

\begin{align}
s_{1}''\textsuperscript{*}=s_1\textsuperscript{*}+\frac{s_{1}'-s_1}{x_{1}\textsuperscript{2}+x_{2}\textsuperscript{2}}(x_{1}x_{1}\textsuperscript{*}+x_{2}x_{2}\textsuperscript{*})+\frac{s_{1}''\textsuperscript{@}-s_1'\textsuperscript{@}}{x_{1}\textsuperscript{@2}+x_{2}\textsuperscript{@2}}(x_{1}\textsuperscript{@}x_{1}\textsuperscript{*}+x_{2}\textsuperscript{@}x_{2}\textsuperscript{*})
\label{eq:12}
\end{align}
\begin{align*}
s_{1}'''\textsuperscript{\#}=s_1\textsuperscript{\#}+\frac{s_{1}'-s_1}{x_{1}\textsuperscript{2}+x_{2}\textsuperscript{2}}(x_{1}x_{1}\textsuperscript{\#}+x_{2}x_{2}\textsuperscript{\#})+\frac{s_{1}''\textsuperscript{@}-s_1'\textsuperscript{@}}{x_{1}\textsuperscript{@2}+x_{2}\textsuperscript{@2}}(x_{1}\textsuperscript{@}x_{1}\textsuperscript{\#}+x_{2}\textsuperscript{@}x_{2}\textsuperscript{\#})+\\
\frac{s_{1}'''\textsuperscript{*}-s_1''\textsuperscript{*}}{x_{1}\textsuperscript{*2}+x_{2}\textsuperscript{*2}}(x_{1}\textsuperscript{*}x_{1}\textsuperscript{\#}+x_{2}\textsuperscript{*}x_{2}\textsuperscript{\#})
\end{align*}
\begin{align*}
s_{2}''\textsuperscript{*}=s_2\textsuperscript{*}+\frac{s_{2}'-s_2}{x_{1}\textsuperscript{2}+x_{2}\textsuperscript{2}}(x_{1}x_{1}\textsuperscript{*}+x_{2}x_{2}\textsuperscript{*})+\frac{s_{2}''\textsuperscript{@}-s_2'\textsuperscript{@}}{x_{1}\textsuperscript{@2}+x_{2}\textsuperscript{@2}}(x_{1}\textsuperscript{@}x_{1}\textsuperscript{*}+x_{2}\textsuperscript{@}x_{2}\textsuperscript{*})
\end{align*}
\begin{align*}
s_{2}'''\textsuperscript{\#}=s_2\textsuperscript{\#}+\frac{s_{2}'-s_2}{x_{1}\textsuperscript{2}+x_{2}\textsuperscript{2}}(x_{1}x_{1}\textsuperscript{\#}+x_{2}x_{2}\textsuperscript{\#})+\frac{s_{2}'\textsuperscript{@}'-s_2'\textsuperscript{@}}{x_{1}\textsuperscript{@2}+x_{2}\textsuperscript{@2}}(x_{1}\textsuperscript{@}x_{1}\textsuperscript{\#}+x_{2}\textsuperscript{@}x_{2}\textsuperscript{\#})+\\
\frac{s_{2}'''\textsuperscript{*}-s_2''\textsuperscript{*}}{x_{1}\textsuperscript{*2}+x_{2}\textsuperscript{*2}}(x_{1}\textsuperscript{*}x_{1}\textsuperscript{\#}+x_{2}\textsuperscript{*}x_{2}\textsuperscript{\#})
\end{align*}

If $w_1,w_2,w_3,w_4$ are not updated, then non-linear `compensation weights' of $p$, $q$ must be added to $w_5,w_6$ respectively in order to preserve $Y$ output updation. To derive values of $p,q$, consider that for the initial updation, using equations \ref{eq:1}, \ref{eq:2}, \ref{eq:3}, \ref{eq:5} we can say that:
\begin{align}
Y_{backpropgation}&=Y_{contributionfactor}
\implies v_1.w_5+v_2.w_6&=v_1^c.(w_5+p)+v_2^c.(w_6+q)
\label{eq:13}
\end{align}
where $v_1^c$ and $v_2^c$ are the constants-- their values don't change, as $w_1$,$w_2$,$w_3$,$w_4$ don't get updated.
So, for the first updation, we get:
\begin{align*}
v_{1}'.w_{5}'+v_{2}'.w_{6}'&=v_1^c.(w_{5}'+p)+v_2^c.(w_{6}'+q)\\
\implies (v_1^c+dv_{1}').w_{5}'+(v_2^c+dv_{2}').w_{6}'&=v_1^c.(w_{5}'+p)+v_2^c.(w_{6}'+q)
\end{align*}
where $w_5',w_6'$ are updated as per equation \ref{eq:5}. Now, compensating $v_1$ and $v_2$ separately, we have:
\begin{align}
(v_1^c+dv_{1}').w_{5}'&=v_1^c.(w_{5}'+p) \implies dv_{1}'.w_{5}'&=v_1^c.p
\label{eq:14}
\end{align}
Similarly,
\begin{align*}
(v_2^c+dv_{2}').w_{6}'&=v_2^c.(w_{6}'+q) \implies dv_{2}'.w_{6}'&=v_2^c.q
\end{align*}
During updation between the 0\textsuperscript{th} and 1\textsuperscript{st} iterations, $dv_1$' can have 3 possible values: (i) RELU is dead at the  0\textsuperscript{th} iteration and stays dead after the 1\textsuperscript{st} update, (ii) RELU is active at the 0\textsuperscript{th} iteration and stays active after the  1\textsuperscript{st} update, and (iii) RELU is dead at the  0\textsuperscript{th} iteration and becomes active after the 1\textsuperscript{st} update:
\begin{align}
dv_{1}'&=0
\label{eq:15}
\end{align}
\begin{align*}
dv_{1}'&=dw_{1}'.x_1+dw_{2}'.x_2 \hspace{0.2cm} ((s_{1}\geq0)\&\&(s_{1}'\geq0))\\
dv_{1}'&=-v_{1}=-s_{1} \hspace{0.2cm} ((s_{1}\geq0)\&\&(s_{1}'<0))
\end{align*}
where
\begin{align*}
s_{1}&=w_{1}.x_1+w_{2}.x_2\\
s_{1}'&=s_{1}(s_{1}<0)\\
\end{align*}
i.e.,
\begin{align}
s_{1}'&=(w_{1}'.x_1+w_{2}'.x_2)(s_{1}\geq0)
\label{eq:16}
\end{align}
\begin{align*}
\implies s_{1}'&=(w_{1}-\eta\frac{dE}{dw_1}).x_1+(w_{2}-\eta\frac{dE}{dw_2}).x_2\\
\implies s_{1}'&=(w_{1}-\eta\frac{dE}{dY}.w_{5}'.x_1(s_{1}\geq0)).x_1+(w_{2}-\eta\frac{dE}{dY}.w_{5}'.x_2(s_{1}\geq0)).x_2\\
\implies s_{1}'&=s_{1}-\eta\frac{dE}{dY}.w_{5}'(x_1^2+x_2^2)(s_{1}\geq0)
\end{align*}

Using equation \ref{eq:16} in \ref{eq:15},

\begin{align}
dv_{1}'&=-\eta\frac{dE}{dY}.w_{5}'(x_1^2+x_2^2) (((w_{1}.x_1+w_{2}.x_2)\geq0)\&\&\\\nonumber
&((w_{1}-\eta\frac{dE}{dY}.w_{5}'.x_1).x_1+(w_{2}-\eta\frac{dE}{dY}.w_{5}'.x_2).x_2))\geq0))
\label{eq:17}
\end{align}
\begin{align*}
dv_{1}'&=-(w_{1}.x_1+w_{2}.x_2) (((w_{1}.x_1+w_{2}.x_2)\geq0)\&\&\\\nonumber
&((w_{1}-\eta\frac{dE}{dY}.w_{5}'.x_1).x_1+(w_{2}-\eta\frac{dE}{dY}.w_{5}'.x_2).x_2))<0))
\end{align*}

Using equation \ref{eq:14} and \ref{eq:17}, we get:
\begin{align}
\scriptstyle
p&=\scriptstyle
\frac{dv_{1}'.w_{5}'}{v_1^c}\\
\scriptstyle
&=\scriptstyle
\frac{-\eta\frac{dE}{dY}.w_{5}'(x_1^2+x_2^2) \scriptstyle
(((w_{1}.x_1+w_{2}.x_2)\geq0)\&\&((w_{1}-\eta\frac{dE}{dY}.w_{5}'.x_1).x_1+(w_{2}-\eta\frac{dE}{dY}.w_{5}'.x_2).x_2))\geq0)).w_{5}'}{v_1^c}\nonumber\\
\scriptstyle
&=\scriptstyle
\frac{-\eta\frac{dE}{dY}.w_{5}'^2(x_1^2+x_2^2) (((w_{1}.x_1+w_{2}.x_2)\geq0)\&\&((w_{1}-\eta\frac{dE}{dY}.w_{5}'.x_1).x_1+(w_{2}-\eta\frac{dE}{dY}.w_{5}'.x_2).x_2))\geq0))}{v_1^c}\nonumber
\label{eq:18}
\end{align}

\begin{align}
\scriptstyle
p&=\scriptstyle
\frac{dv_{1}'.w_{5}'}{v_1^c}\\
\scriptstyle
&=\scriptstyle
\frac{-(w_{1}.x_1+w_{2}.x_2) (((w_{1}.x_1+w_{2}.x_2)\geq0)\&\&((w_{1}-\eta\frac{dE}{dY}.w_{5}'.x_1).x_1+(w_{2}-\eta\frac{dE}{dY}.w_{5}'.x_2).x_2))\geq0)).w_{5}'}{v_1^c}\nonumber\\
\scriptstyle
&=\scriptstyle
\frac{-v_1^c.(((w_{1}.x_1+w_{2}.x_2)\geq0)\&\&((w_{1}-\eta\frac{dE}{dY}.w_{5}'.x_1).x_1+(w_{2}-\eta\frac{dE}{dY}.w_{5}'.x_2).x_2))\geq0)).w_{5}'}{v_1^c}\nonumber\\
\scriptstyle
&=\scriptstyle
-w_{5}'.(((w_{1}.x_1+w_{2}.x_2)\geq0)\&\&((w_{1}-\eta\frac{dE}{dY}.w_{5}'.x_1).x_1+(w_{2}-\eta\frac{dE}{dY}.w_{5}'.x_2).x_2))\geq0))\nonumber
\label{eq:19}
\end{align}

The above represents the amount of weight $p$ to be added for the given iteration. Consider only a single input scheme, i.e., [$x_1$,$x_2$], such that additional compensation to $w_5'$ at each iteration is:

\begin{align*}
\hspace{0.3cm} {\frac{dE}{dw_5'}}=p
\end{align*}

Then using the chain rule, we can generalize this to calculate the total compensation weight $P$ that must be added at the $n_{th}$ iteration as follows:

\begin{align}
Additional \hspace{0.3cm} -\eta\frac{dE}{dY}.v_1^c.\frac{dP}{dw_5}&=p\\
\implies \frac{dP}{dw_5}&=\frac{-p}{\eta\frac{dE}{dY}.v_1^c}\nonumber\\
\implies P&=\int{\frac{-p}{\eta\frac{dE}{dY}.v_1^c}}.dw_5\nonumber
\label{eq:20}
\end{align}

We can find P in terms of $w_5$' using equation 18, 20 as:

\begin{align}
\scriptstyle
&=\scriptstyle
\int{\frac{\frac{\eta\frac{dE}{dY}.w_{5}'^2(x_1^2+x_2^2) (((w_{1}.x_1+w_{2}.x_2)\geq0)\&\&((w_{1}-\eta\frac{dE}{dY}.w_{5}'.x_1).x_1+(w_{2}-\eta\frac{dE}{dY}.w_{5}'.x_2).x_2))\geq0))}{v_1^c}}{\eta\frac{dE}{dY}.v_1^c.v_1^c}}.dw_{5}'\\
\scriptstyle
&=\scriptstyle
\int{\frac{w_{5}'^2(x_1^2+x_2^2) (((w_{1}.x_1+w_{2}.x_2)\geq0)\&\&((w_{1}-\eta\frac{dE}{dY}.w_{5}'.x_1).x_1+(w_{2}-\eta\frac{dE}{dY}.w_{5}'.x_2).x_2))\geq0))}{v_1^{c^2}}}.dw_{5}'\nonumber\\
\scriptstyle
&=\scriptstyle
\frac{w_{5}'^3(x_1^2+x_2^2) (((w_{1}.x_1+w_{2}.x_2)\geq0)\&\&((w_{1}-\eta\frac{dE}{dY}.w_{5}'.x_1).x_1+(w_{2}-\eta\frac{dE}{dY}.w_{5}'.x_2).x_2))\geq0))}{3v_1^{c^2}}\nonumber
\label{eq:21}
\end{align}

and using equation 19, 20 as:

\begin{align}
\scriptstyle
&=\scriptstyle
\int{\frac{w_{5}' (((w_{1}.x_1+w_{2}.x_2)\geq0)\&\&((w_{1}-\eta\frac{dE}{dY}.w_{5}'.x_1).x_1+(w_{2}-\eta\frac{dE}{dY}.w_{5}'.x_2).x_2))\geq0))}{\eta\frac{dE}{dY}.v_1^c}}.dw_{5}'\\
\scriptstyle
&=\scriptstyle
\frac{w_{5}'^2 (((w_{1}.x_1+w_{2}.x_2)\geq0)\&\&((w_{1}-\eta\frac{dE}{dY}.w_{5}'.x_1).x_1+(w_{2}-\eta\frac{dE}{dY}.w_{5}'.x_2).x_2))\geq0))}{2\eta\frac{dE}{dY}.v_1^{c}}\nonumber
\label{eq:22}
\end{align}

Here, $p$ and $P$ compensate for weights $w_1,w_2$. Similarly, $q$ and $Q$ can be defined to compensate for weights $w_3,w_4$.  Let $w_5+p$ be $r_5$ and $w_6+q$ be $r_6$. $r$ may be interpreted as a transformation on $w_{layer2}$ that takes care of the contribution factor from $w_{layer1}$. Hence, the general equation for updation after `n' iterations is:

\begin{align*}
 r_{5n}&=w_{5n}+\frac{w_{5n}^3(x_1^2+x_2^2)}{3(v_{1c}+\sum_1^n \frac{dE}{dY}_{n} w_{5n}(x_1^2+x_2^2)^2)}\hspace{0.2cm} (A_{n-1}>0\hspace{0.2cm}\&\& \hspace{0.2cm}A_{n}>0)\\
   &=w_{5n}+\frac{w_{5n}^2}{2\eta\frac{dE}{dY}_{n}(v_{1c}+\sum_1^n \frac{dE}{dY}_{n} w_{5n}(x_1^2+x_2^2))}\hspace{0.2cm}(A_{n-1}>0\hspace{0.2cm}\&\&\hspace{0.2cm}A_{n}<0)\\
\end{align*}

where $A_n=(w_{1c}+\sum_1^n \frac{dE}{dY}_{n} w_{5n}x_1)x_1+(w_{2c}+\sum_1^n \frac{dE}{dY}_{n} w_{5n}x_2)x_2$\\\\

\begin{align*}
 r_{6n} &=w_{6n}+\frac{w_{6n}^3(x_1^2+x_2^2)}{3(v_{2c}+\sum_1^n \frac{dE}{dY}_{n} w_{6n}(x_1^2+x_2^2)^2)}\hspace{0.2cm} (B_{n-1}>0\hspace{0.2cm}\&\&\hspace{0.2cm}B_{n}>0)\\
   &=w_{6n}+\frac{w_{6n}^2}{2\eta\frac{dE}{dY}_{n}(v_{2c}+\sum_1^n \frac{dE}{dY}_{n} w_{6n}(x_1^2+x_2^2))}\hspace{0.2cm}(B_{n-1}>0\hspace{0.2cm}\&\&\hspace{0.2cm}B_{n}<0)\\
\end{align*}

where $B_n=(w_{3c}+\sum_1^n  \frac{dE}{dY}_n w_{6n}x_1)x_1+(w_{4c}+\sum_1^n  \frac{dE}{dY}_n w_{6n}x_2)x_2$\\

Here subscript $n$ represents value at the $n^{th}$ iteration, and subscript $c$ represents the constant value, i.e., the initial value which does not change across iterations.

In the condition $s_1\geq0$ and $s_1'\geq0$, using equations \ref{eq:5}, \ref{eq:13}, 18, for inital input [$x_1$,$x_2$] we have:
\begin{align*}
    &p=-\eta\frac{dE}{dy}w_5'\textsuperscript{2}\frac{(x_1\textsuperscript{2}+x_2\textsuperscript{2})}{v_1\textsuperscript{c}},\
        q=-\eta\frac{dE}{dy}w_6'\textsuperscript{2}\frac{(x_1\textsuperscript{2}+x_2\textsuperscript{2})}{v_2\textsuperscript{c}}\nonumber\\
    &y'=(w_5'+p)v_1\textsuperscript{c}+(w_6'+p)v_2\textsuperscript{c},\ \frac{dE}{dy}'=y'-y_g\nonumber\\
    &w_5''=w_5'-\eta \frac{dE}{dy}'v_1'=w_5'-\eta(v_1\textsuperscript{c} w_5'+v_1\textsuperscript{c}p+v_2\textsuperscript{c} w_6'+v_2\textsuperscript{c} q-y_g)(v_1\textsuperscript{c}-pv_1\textsuperscript{c})\nonumber\\
       & w_6''=w_6'-\eta \frac{dE}{dy}'v_1'=w_5'-\eta(v_1\textsuperscript{c} w_5'+v_1\textsuperscript{c}p+v_2\textsuperscript{c} w_6'+v_2\textsuperscript{c} q-y_g)(v_2\textsuperscript{c}-qv_2\textsuperscript{c})
\end{align*}

For next input [$x_1$\textsuperscript{@},$x_2$\textsuperscript{@}] we have:
\begin{align}
    &v_1\textsuperscript{@}'w_5'=(w_5'+p+p\textsuperscript{@})v_1\textsuperscript{@}\nonumber\\
    &\implies -\eta \frac{dE}{dy}w_5'\textsuperscript{2}(x_1x_1\textsuperscript{@}+x_2x_2\textsuperscript{@})=p\textsuperscript{@}(w_1x_1\textsuperscript{@}+w_2x_2\textsuperscript{@})+p(w_1x_1\textsuperscript{@}+w_2x_2\textsuperscript{@})\nonumber\\
    &p\textsuperscript{@}=p- \eta \frac{dE}{dy}w_5'\textsuperscript{2}\frac{(x_1x_1\textsuperscript{@}+x_2x_2\textsuperscript{@})}{v_1'\textsuperscript{@}}\nonumber\\
    &\text{Parallely,}\  q\textsuperscript{@}=q- \eta \frac{dE}{dy}w_6'\textsuperscript{2}\frac{(x_1x_1\textsuperscript{@}+x_2x_2\textsuperscript{@})}{v_2'\textsuperscript{@}}
\end{align}
From the above, we see that for multiple inputs, the entire input sequence must be stored to calculate the $P$ and $Q$ values. Therefore, in the cases of updating either (i) $w_5,w_6$ by adding compensatory weights $p,q$, or (ii) updating the compensatory weights $p,q$ for a fixed $w_5, w_6$, we only use one input.

To handle multiple inputs, we update $s_1$, $s_2$ as follows:

\begin{align}
\text{Initially:}&\nonumber\\
    &s_1=w_1 x_1+w_2 x_2,\ s_2=w_3 x_1+w_4 x_2\nonumber\\
    &S_1=w_1 x_1-w_2 x_2,\ S_2=w_3 x_1-w_4 x_2\nonumber\\
\text{Updated as:}&\nonumber\\
    &s_1\textsuperscript{@}'=\frac{s_1'+S_1'}{2}x_1\textsuperscript{@}+\frac{s_1'-S_1'}{2}x_2\textsuperscript{@},\
    s_2\textsuperscript{@}'=\frac{s_2'+S_2'}{2}x_1\textsuperscript{@}+\frac{s_2'-S_2'}{2}x_2\textsuperscript{@}\nonumber\\
    &S_1\textsuperscript{@}'=\frac{s_1'+S_1'}{2}x_1\textsuperscript{@}-\frac{s_1'-S_1'}{2}x_2\textsuperscript{@},\
    S_2\textsuperscript{@}'=\frac{s_2'+S_2'}{2}x_1\textsuperscript{@}-\frac{s_2'-S_2'}{2}x_2\textsuperscript{@}
\end{align}

Above, we have shown that in the context of \ref{deq1}, ($W2+p$) (where $W_2$ is the second weight layer comprising of $w_5, w_6$) can contribute (attend for text) equivalent to updated $W_1$, and $W_2$. $p$ is the contribution/weight attention factor of $W_1$ to $W_2$, i.e., the amount by which $W_1$ contributes/attends to $W_2$. In this case, $W_2$ weights are now non-linear-- based on equation \ref{eq:21}, $W_2$ weights can take the form of $W_2+W_2^3$. By introducing non-linear weights, we can therefore reduce the total number of weights by 1 set (i.e., the number of weights corresponding to a particular layer). By showing that NN of n layers can be collapsed to n-1 layers, we can further use the method of induction to theoretically prove that NN can be collapsed from $(n-1)->(n-2)->...->1$ layer, such that the final network has just non-linear weights for layer $W_n$.

This method of $W_n$ calculation is the \textit{Front Contribution Algorithm}, as instead of propagating error backwards, the network propagates contribution forward to collapse the network. \\

\textbf{Experiments:}

\begin{figure*}[ht!]
    \centering
    \includegraphics[width=0.5\textwidth]{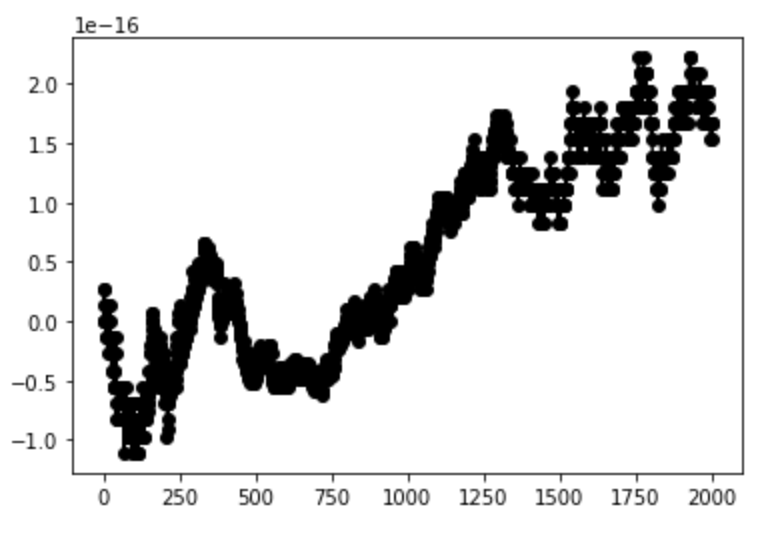}
    \includegraphics[width=0.5\textwidth]{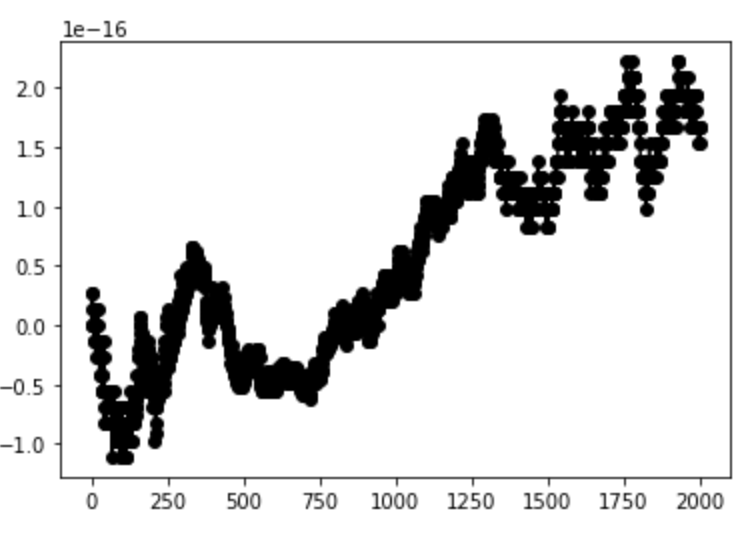}
    \includegraphics[width=0.5\textwidth]{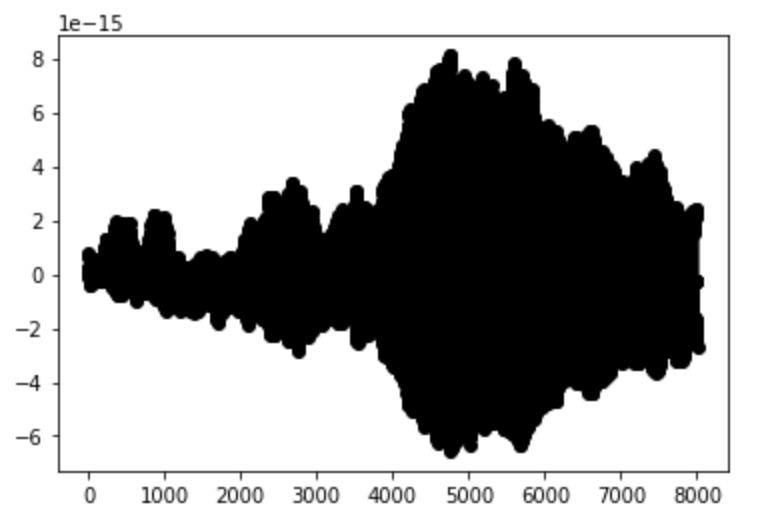}
    \caption{Here in all 3 figures,  x-axis represents the training iteration, and the y-axis represents the error between output produced through backpropagation and the ouptut produced by forward contribution. Top: On updation of weights $w_5, w_6$ as $w_5+p, w_6+q$. Middle: On updation of values of $p, q$ without updation of $w_5, w_6$. Bottom: Updation of states $v_1, v_2$ without weight updation.}
    \label{fig:res}
\end{figure*}

The results of the three techniques we use for Forward Contribution, over the implementation of the XOR task are displayed in Figure \ref{fig:res}.

The order of error is $<10^{-15}$, proving that Forward Contribution is a true equivalent of backpropagation.
 

\begin{figure*}[t]
    \centering
    \includegraphics[width=\textwidth]{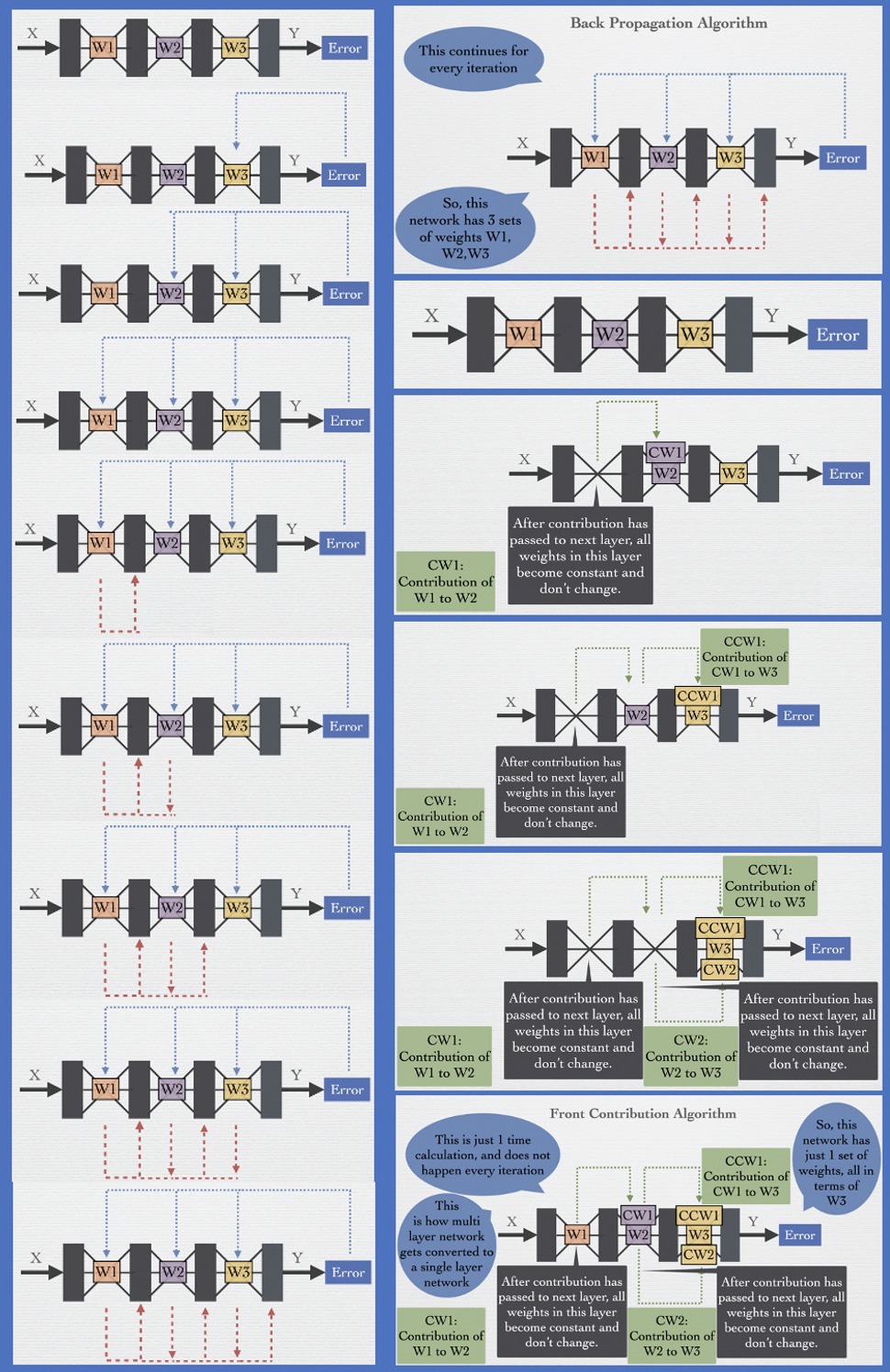}
    \caption{On the left hand side, from top->bottom are the steps for Back-Propagation; the top right image has the final summary of Back-Propagation. The remaining images on the right show how Front-Contribution functions (top->bottom).}
    \label{fig:illu}
\end{figure*}

\textbf{Backpropagation vs. Front Contribution Illustrations:}
We illustrate the process flows of Back Propagation and Front Contribution in Figure \ref{fig:illu}.

\end{document}